\documentclass[11pt,a4paper]{article}
\usepackage[hyperref]{acl2019}

\usepackage{url}

\usepackage{times}
\usepackage{latexsym}
\usepackage{multirow}
\usepackage{url}

\usepackage{times}
\usepackage{amsmath}
\usepackage{graphicx}
\usepackage{arabtex}
\usepackage[T1]{fontenc}
\usepackage{wrapfig}
\usepackage{enumitem}
\usepackage{colortbl,xcolor}
\usepackage{footnote}

\usepackage{tipa}

\newcommand{\hide}[1]{}

\usepackage{float}

\aclfinalcopy % Uncomment this line for the final submission
\title{Adversarial Multitask Learning \\ for Joint Multi-Feature and Multi-Dialect Morphological Modeling }

\author{Nasser Zalmout \and Nizar Habash \\
  Computational Approaches to Modeling Language Lab \\
  New York University Abu Dhabi \\
  United Arab Emirates \\
  {\tt \{nasser.zalmout,nizar.habash\}@nyu.edu} \\}

\date{}

\begin{document}
\maketitle

\setarab
\novocalize

\begin{abstract}

Morphological tagging is challenging for morphologically rich languages due to the large target space and the need for more training data to minimize model sparsity. Dialectal variants of morphologically rich languages suffer more as they tend to be more noisy and have less resources. In this paper we explore the use of multitask learning and adversarial training to address morphological richness and dialectal variations in the context of full morphological tagging. We use multitask learning for joint morphological modeling for the features within two dialects, and as a knowledge-transfer scheme for cross-dialectal modeling. We use adversarial training to learn dialect invariant features that can help the knowledge-transfer scheme from the high to low-resource variants. 
We work with two dialectal variants: Modern Standard Arabic (high-resource ``dialect''\footnote{We view Arabic as a collective of dialectal variants in which MSA is the main high-resource dialect,
and EGY is a low-resource dialect. We therefore use ``variant'' and ``dialect'' interchangeably. }) and Egyptian Arabic (low-resource dialect) as a case study. Our models achieve state-of-the-art results for both. Furthermore, adversarial training provides more significant improvement when using smaller training datasets in particular.

\end{list}
\end{abstract}

\section{Introduction}
Morphological tagging for morphologically rich languages (MRL) involves modeling interdependent features, with a large combined target space. Joint modeling of the different features, through feature concatenation, results in a large target space with increased sparsity. Whereas total separation of the different feature models eliminates access to the other features, which constrains the model. These issues are further exacerbated for dialectal content, with many morphosyntactic variations that further complicate the modeling.

In this paper we work with Modern Standard Arabic (MSA) and Egyptian Arabic (\textsc{Egy}), both MRLs, and dialectal variants. Written Arabic text is also highly ambiguous, due to its diacritic-optional orthography, resulting in several interpretations of the same surface forms, and further increasing sparsity. Joint modeling is particularly promising for such ambiguous nature as it supports identifying more complex patterns involving multiple features. In \textsc{Egy}, for example, the suffix <nA> {\it nA}
`we, us, our'  in the word <drsnA> {\it drsnA} can be the subject of the perfective 1st person plural verb (`we studied'), the 1st person plural object clitic of a perfective 3rd person masculine singular verb (`he taught us'), or the 1st person plural possessive pronoun for the nominal (`our lesson'), among other possible interpretations.

Morphological tagging models rely heavily on the availability of large annotated training datasets. Unlike \textsc{MSA}, Arabic Dialects are generally low on resources. In this paper we also experiment with knowledge-transfer models from high to low-resource variants. The similarities between the Arabic variants, both for \textsc{MSA} and Dialectal Arabic (\textsc{DA}), like \textsc{Egy}, should facilitate knowledge-transfer, making use of the resources of the high-resource variants. We use multitask learning architectures in several configurations for cross-dialectal modeling. 
We further investigate the best approaches and configurations to use word and character embeddings in the cross-dialectal multitask learning model, and whether mapping the various pretrained word embedding spaces is beneficial.  Despite having several contributions in the literature, the role of mapped embedding spaces has not been studied in the context of joint morphological modeling of different dialects.

Finally, we use adversarial training to learn dialect-invariant features for MSA and \textsc{Egy}. The intuition is to make the modeling spaces for both variants closer to each other, which should facilitate the knowledge-transfer scheme from the high-resource (MSA) to the low-resource (\textsc{Egy}) sides. 

Our models achieve state-of-the-art morphological disambiguation results for both MSA and \textsc{Egy}, with up to 10\% relative error reduction. Adversarial training proved more useful when using a smaller \textsc{Egy} training datasets in particular, simulating lower-resource settings.
The contributions of the paper include (1) a joint multi-feature and cross-dialectal morphological disambiguation model for several MRL variants, (2) adversarial training for cross-dialectal morphological knowledge-transfer.

\section{Linguistic Motivation}
MRLs, like Arabic, have many morphemes that represent several morphological features. The target space for the combined morphological features in MRLs therefore tends to be very large. MRLs also tend to have more inflected words than other languages. 
MRLs also usually have a higher degree of ambiguity, with different interpretations of the same surface form. In Arabic, this ambiguity is exacerbated by the diacritization-optional orthography, which results in having about 12 analyses per word on average \cite{Habash_10_book}.

One approach to model morphological richness and ambiguity is to use \textit{morphological analyzers}, which are used to encode all potential word inflections in the language. The ideal morphological analyzer should return all the possible analyses of a surface word (modeling ambiguity), and cover all the inflected forms of a word lemma (modeling morphological richness). The best analysis is then chosen through \textit{morphological disambiguation}, which is essentially part-of-speech tagging for all the features in addition to lemma and diacritized form choices.

MSA is the written Arabic that is mainly used in formal settings. \textsc{DA}, like \textsc{Egy}, on the other hand, is the primarily spoken language used by native Arabic speakers in daily exchanges. \textsc{DA} has recently seen an increase in written content, due to the growing social media use in the region.
\textsc{DA}, similar to MSA, is also morphologically rich, with a high degree of ambiguity. \textsc{DA} spans many Arabic dialects that are used across the Arab World, and they vary by the regions and cities they are used in \cite{Bouamor-etal-2018}. The large number of \textsc{DA} variants, along with it being mainly spoken, result in \textsc{DA} being usually low on resources. 

\textsc{MSA} and DA have many morphological, lexical and syntactic similarities that a cross-dialectal model can leverage  \cite{CALIMA:2012}. DA has many \textsc{MSA} cognates, both \textsc{MSA} and DA use the same script, and DA content in general includes a lot of code-switching with MSA.\footnote{Although \textsc{Egy}, like DA in general, does not have a standardized orthography like \textsc{MSA} \cite{habash-etal-2018-unified}.} These similarities can be useful in a joint learning model, enabling a knowledge-transfer scheme, especially from the high-resource to low-resource variants.

In this paper we focus on \textsc{Egy} as an example of DA. The set of morphological features that we model for both \textsc{MSA} and \textsc{Egy} can be:

\begin{itemize}
\item Open-Set Features: Lemmas (lex) and diacritized forms (diac), henceforth "lexicalized features". These features are unrestricted and have large and open vocabularies.
\item Closed-Set Features: A set of 14 features, including inflectional features and clitics, each with a corresponding set of values/tags that are predicted using taggers. The inflectional features include: part-of-speech (POS), aspect (asp), case (cas), gender (gen), person (per), number (num), mood (mod), state (stt), voice (vox). The clitics include: enclitics, like pronominal and negative particle enclitics; proclitics, like article proclitic, preposition proclitics, conjunction proclitics, question proclitics.

\end{itemize}

Morphological disambiguation involves predicting the values for each of these features, then using these predictions to rank the different analyses from the morphological analyzer.

\section{Background and Related Work}
\paragraph{Joint Modeling in NLP} Joint NLP modeling in general has been an active area of research throughout the past several years, supported by recent updates in deep learning architectures. Multitask learning models have been proven very useful for several NLP tasks and applications, \cite{collobert2011natural,Sogaard_16_ACLshort,Alonso_17_EACL,bingel:2017,Hashimoto_2017_EMNLP}. \newcite{Inoue-CONLL-2017} used multitask learning for fine-grained POS tagging in \textsc{MSA}. We extend their work by doing cross-dialectal modeling and various contributions for low-resource dialects.

\paragraph{Cross-Lingual Transfer} Cross-lingual morphology and syntax modeling has also been a very active NLP research area, with contributions in morphological reinflection and paradigm completion \cite{aharoni2016improving,faruqui2016morphological,kann2017one}, morphological tagging \cite{buys2016cross,Cotterell_17_EMNLP}, parsing \cite{guo2015cross,ammar2016many},  among others. \newcite{Cotterell_17_EMNLP} used multitask learning for multi-lingual POS tagging, similar in spirit to our approach. Their architecture, however, models the morphological features in each language in a single task, where each target value represents all morphological features combined. This architecture is not suitable for MRLs, with large target spaces.

\paragraph{Adversarial Domain Adaptation} Inspired by the work of \newcite{goodfellow:2014}, adversarial networks have been used to learn domain invariant features in models involving multiple domains, through domain adversarial training \cite{ganin2015,ganin2016}. Adversarial training facilitates domain-adaptation schemes, especially in high-resource to low-resource adaptation scenarios. The approach is based on an adversarial discriminator, which tries to identify the domain of the data, and backpropagates the negative gradients in the backward direction. This enables the model to learn shared domain features. Adversarial domain adaptation has been used in several NLP applications, including sentiment analysis \cite{chen2016}, POS tagging for Twitter \cite{gui2017}, relation extraction \cite{fu2017,wang2018}, among other applications. As far as we know, we are the first to apply adversarial domain adaptation in the context of dialectal morphological modeling.

\paragraph{Arabic Morphological Modeling} Morphological modeling for Arabic has many contributions in both \textsc{MSA} \cite{Diab:2004a,ACL:habash-rambow:2005,MADAMIRA:2014,FARASA:2016NAACL,khalifayamama}, and Dialectal Arabic  \cite{Duh:2005,konvens:09_alsabbagh12o,Habash_13_NAACL}. There were also several neural extensions that show impressive results \cite{Zalmout-2017-EMNLP,Zalmout_2018_NAACL}. These contributions use separate models for each morphological feature, then apply a disambiguation step, similar to several previous models for Arabic \cite{ACL:habash-rambow:2005,MADAMIRA:2014}. \newcite{Shen:2016} use LSTMs with word/character embeddings for Arabic tagging. \newcite{Darwish:2018} use a CRF model for a multi-dialect POS tagging, using a small annotated Twitter corpus.  \newcite{Alharbi:2018} also use neural models for Gulf Arabic, with good results.

\section{Baseline Tagging and Disambiguation Architecture}
In this section we present our baseline tagging and disambiguation architectures. We extend this architecture for joint modeling in the section that follows.

\subsection{Morphological Feature Tagging}
We use a similar tagging architecture  to \newcite{Zalmout_2018_NAACL}, based on a Bi-LSTM tagging model, for the closed-set morphological features.
Given a sentence of length L  \(\{w_1, w_2, ..., w_L\)\}, every word \(w_j\) is represented by vector \(\mathbf{v}_j\). We use two LSTM layers to model the relevant context for each direction of the target word, using:
\[\overrightarrow{\hat{\mathbf{h}}}_j=g(\mathbf{v}_j,\overrightarrow{\mathbf{h}}_{j-1})\]
\[\overleftarrow{\hat{\mathbf{h}}}_j=g(\mathbf{v}_j,\overleftarrow{\mathbf{h}}_{j+1})\]
where \(\mathbf{h}_j\) is the context vector from the LSTM for each direction. We join both sides, apply a non-linearity function, output layer, and softmax for a probability distribution. The input vector \(\mathbf{v}_j\) is comprised of:
\[\mathbf{v}_j = [\mathbf{w}_j; \mathbf{s}_{j}; \mathbf{a}_j^{f}]\]

Where $\mathbf{w}_j$ is the word embedding vector, \(\mathbf{s}_j\) is a vector representation of the characters within the word, and \(\mathbf{a}_j^{f}\) is a vector representing all the candidate morphological tags (from an analyzer), for feature \(f\).

\begin{figure}[h]
\centerline{%
\includegraphics[scale=0.52]{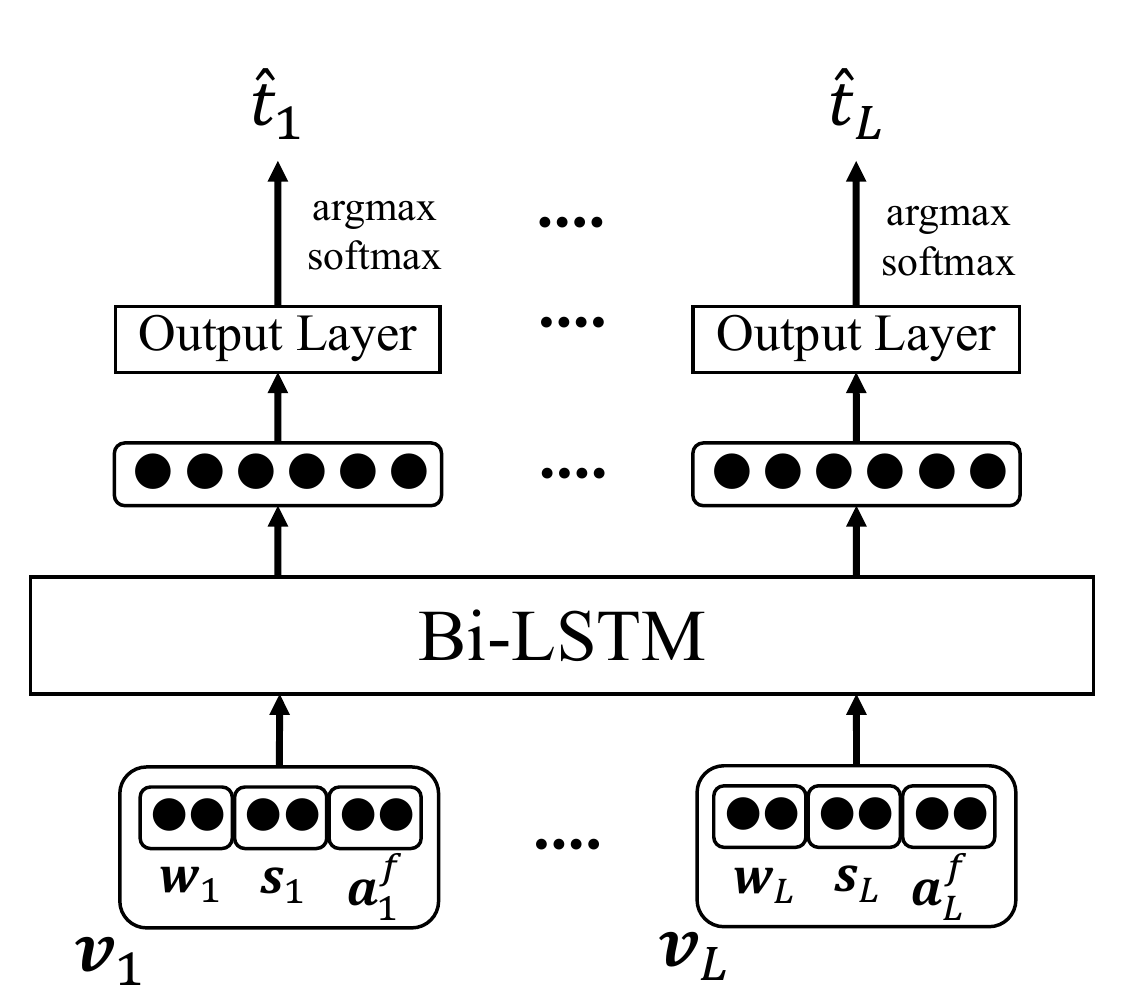}
}%
\caption{The overall tagging architecture, with the input vector as the concatenation of the word, characters, and candidate tag embeddings.}
\label{architecture}
\end{figure}

We pre-train the word embeddings with Word2Vec \cite{Mikolov-NIPS-2013}, using a large external dataset. For the character embeddings vector \(\mathbf{s}_j\) we use an LSTM-based architecture, applied to the character sequence in each word separately. We use the last state vector as the embedding representation of the word's characters. 

The morphological feature vector \(\mathbf{a}_j^{f}\) embeds the candidate tags for each feature. We use a morphological analyzer to obtain all possible feature values of the word to be analyzed, embed the values using a feature-specific embedding tensor, then sum all the resulting vectors for each feature:
\[ \mathbf{a}_j^{f} = \sum_{n=1}^{N_f} \mathbf{a}_{j,n}^{f} \]
Where \(N_f\) is the maximum number of possible candidate tags for the word \(j\) (from the analyzer), for feature \(f\). We sum the vectors because the tags are alternatives, and do not constitute a sequence.

The \(\mathbf{a}_j^{f}\) vector does not constitute a hard constraint and can be discarded if a morphological analyzer is not used.
Figure~\ref{architecture} shows the overall tagging architecture.

\subsection{Lemmatization and Diacritization}
The morphological features that are non-lexical, like POS, gender, number, among others, are handled by the model presented so far, using the multitask learning architecture. Lexical features, like lemmas and diacritized forms, on the other hand, are handled with neural language models, as presented by \newcite{Zalmout-2017-EMNLP} and \newcite{Zalmout_2018_NAACL}. The lexical features are more difficult to model jointly with the non-lexical features, as they have large target spaces, and modeling them as classification tasks is not feasible.

\subsection{Full Morphological Disambiguation}
The predicted feature values for each word, whether from the tagger or the language models, can be returned directly if we do not use a morphological analyzer, without an explicit ranking step. If a morphological analyzer is used, the disambiguation system selects the optimal analysis for the word from the set of analyses returned by the morphological analyzer. We use the predicted feature values from the taggers and language models to rank the analyses, and select the analysis with highest number of matched feature values. We also use weighted matching; where instead of assigning ones and zeros for the matched/mismatched features, we use a feature-specific matching weight.
We replicate the morphological disambiguation pipeline presented in earlier contributions \cite{Zalmout-2017-EMNLP,Zalmout_2018_NAACL}, and use the same parameter values and feature weights.

\section{Multitask Learning Architecture}
Most of the previous approaches for morphological tagging in Arabic learn a separate model for each morphological feature, and combine the predicted tags for disambiguation \cite{MADAMIRA:2014,Zalmout-2017-EMNLP,Zalmout_2018_NAACL}. This hard separation eliminates any knowledge sharing among the different features when training and tagging. Joint learning, through parameter sharing in multitask learning, helps prune the space of target values for some morphological features, and reduce sparsity.
The separation of the morphological models is also inefficient in terms of execution complexity. Training 14 different models, and running them all during runtime, is very wasteful in terms of execution time, memory footprint, and disk space.

Multitask learning is particularly useful in tasks with relatively complementary models, and usually involves primary and auxiliary tasks. We use multitask learning for joint training of the various morphological features. We extend the morphological tagging architecture presented at the previous section into a multitask learning model. We learn the different morphological features jointly through sharing the parameters of the hidden layers in the Bi-LSTM network. The input is also shared, through the word and character embeddings. We also use a unified feature-tags vector representation for all features, through concatenating the \(\mathbf{a}_j^{f}\)  vectors for each feature of each word:
\[ \mathbf{a}_j = [\mathbf{a}_j^{pos};...;\mathbf{a}_j^{num};...;\mathbf{a}_j^{vox}] \]

The output layer is separate for each morphological feature, with separate softmax and argmax operations. 
\begin{figure}[h]
\centerline{%
\includegraphics[scale=0.5]{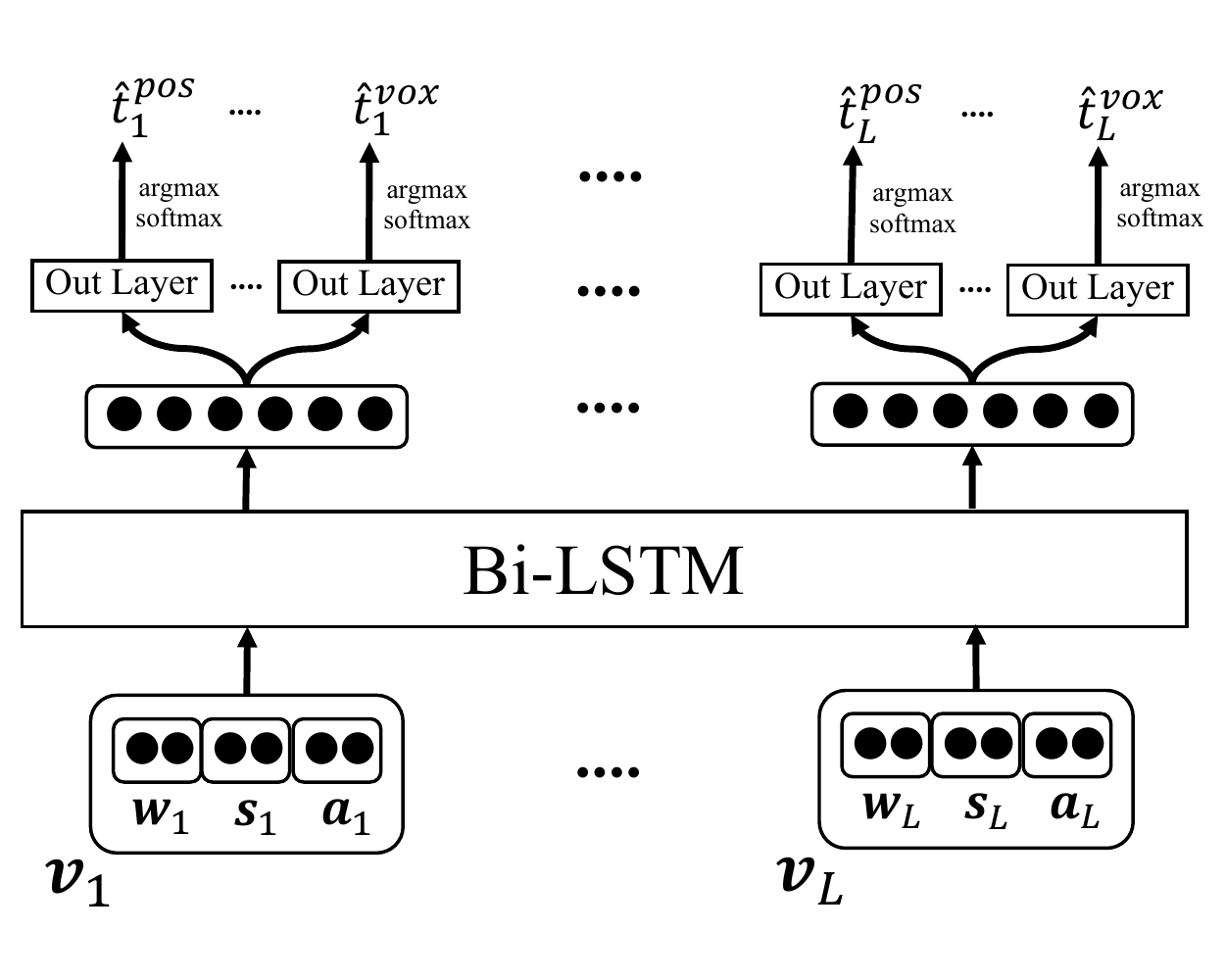}
}%
\caption{The multitask learning architecture, having separate output layers for each feature.}
\label{multitask_architecture}
\end{figure}
The loss function is the average of the individual feature losses, which are based on minimizing cross entropy \(H\) for each feature \(f\):
\[ H(\hat{T}, T) = \frac{1}{|F|}\sum_{f \in F}  H(\hat{t_f}, t_f)  \]

Where \(T\) represents the combined morphological tags for each word, and \(F\) is the set of features \(\{pos,asp,...,vox\}\). Figure~\ref{multitask_architecture} shows the overall architecture for tagging using multitask learning.

\section{Cross-Dialectal Model}
Joint morphological modeling of high-resource and low-resource languages can be very beneficial as a knowledge-transfer scheme. Knowledge-transfer is more viable for languages that share linguistic similarities. In the context of DA, the linguistic similarities between MSA and the dialects, along with the MSA cognates common in DA, should allow for an efficient transfer model. 

We train the model through dividing the datasets of each variant into batches, and running one variant-specific batch at a time. We introduce various extensions to the multitask learning architecture for cross-dialectal modeling. These include sharing the embeddings for the pretrained word embeddings and character embeddings, sharing the output layers for the different features, and adversarial training as a form of dialect adaptation.
The decisions of shared vs joint modeling throughout the various architecture choices  will also affect the size of the model and number of parameters. %In this paper we focus on accuracy, and to less extent the size of the models.

\subsection{Shared Embeddings}
\label{embeddings-section}
Pretrained embeddings have been shown to be very beneficial for several NLP tasks in Arabic  \cite{Zalmout-2017-EMNLP,erdmann-etal-2018-addressing,watson2018utilizing}. In the context of joint modeling of different variants, pretrained embeddings can either be learnt separately or jointly, with several different configurations that include:

\begin{itemize}
  \item Separate embedding spaces, through separate models for the different dialects, trained on separate datasets.
  \item Merged embedding datasets, by merging the datasets for the different dialects and train a single embedding model. This approach is viable because the different Arabic variants use the same script, and DA usually involves a lot of code-switching with MSA.
  \item Mapped embedding spaces, by training separate models for each dialect, then mapping the embedding spaces together.
\end{itemize}
 
We use \textsc{Vecmap} \cite{artetxe2016learning,artetxe2017learning} to map the embedding spaces of the different variants (MSA and DA). \textsc{Vecmap} uses a seed dictionary to learn a mapping function that minimizes the distances between seed dictionary unigram pairs.

In addition to shared word embeddings, the character-level embeddings can also be learned separately or jointly. We do not use pretrained embeddings for the characters, and the embeddings are learnt as part of the end-to-end system.

\subsection{Shared Output Layers}

In the multitask learning architecture, each of the different morphological features needs a separate output layer. In our experiments with Arabic, we are modeling 14 morphological features, which requires 14 output layers. For cross-dialectal modeling, we can have separate output layers for each dialect, which results in 28 output layers for MSA and \textsc{Egy}. Another design choice in this case is to share the output layers between the different dialects, regardless of how many dialects are modeled jointly, with 14 shared output layers only.

Despite the morphological features being similar across the dialects, the target space for each feature might vary slightly for each dialect (as in proclitics and enclitics). 
In the case of shared output layers, we have to merge the target space values for the features of the different dialects, and use this combined set as the target vocabulary.

\subsection{Adversarial Dialect Adaptation}

\begin{figure}[h]
\centerline{%
\includegraphics[scale=0.50]{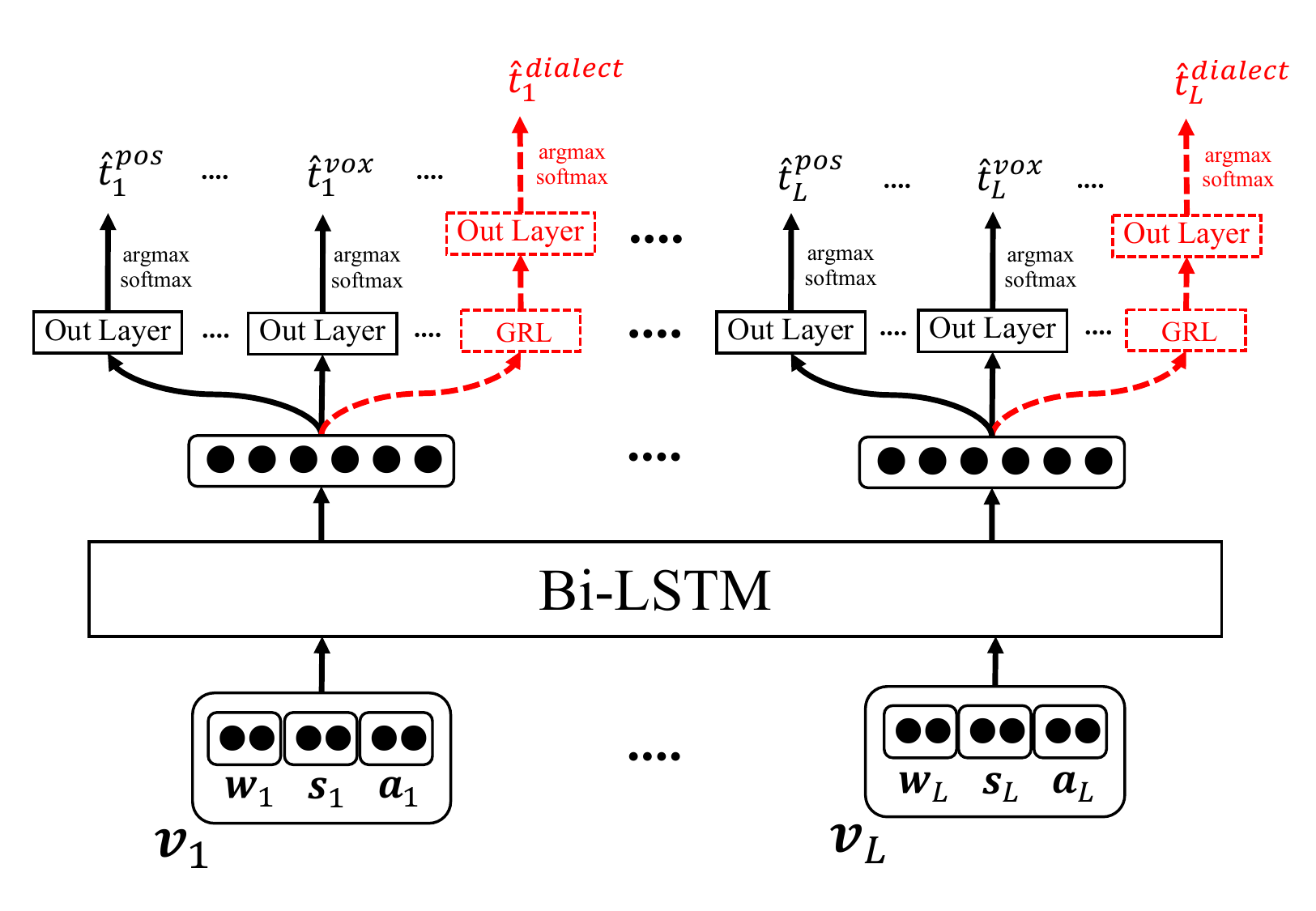}
}%
\caption{The adversarial adaptation architecture, with a discriminator task that backpropagates negative gradients using the Gradient Reversal Layer (GRL).}
\label{adversarial_architecture}
\end{figure}

Similar to adversarial domain adaptation, the goal of the \textit{adversarial dialect adaptation} approach is to learn common features for the different dialects through an adversarial discriminator. Learning dialect-invariant features would facilitate a richer knowledge-transfer scheme from the high-resource to the low-resource variants, since they are both modeled in the same invariant space. Adversarial adaptation can make use of a large annotated dataset from the high-resource dialect,  unlabeled low-resource dialect data, and a small annotated low-resource dialect dataset.
Adversarial adaptation learns dialect invariant features through backpropagating the negative gradients in the backward direction for the discriminator. The backward/forward propagation is managed by the Gradient Reversal Layer. Figure \ref{adversarial_architecture} shows the architecture with the discriminator task.

\paragraph{Gradient Reversal Layer}  Presented by \newcite{ganin2015}, the gradient reversal layer (GRL) passes the identity function in the forward propagation, but negates the gradients it receives in backward propagation, i.e. \(g(F(x))=F(x)\) in forward propagation, but \(\Delta g(F(x))=-\lambda \Delta F(x)\) in backward propagation. \(\lambda\) is a weight parameter for the negative gradient, which can have an update schedule. \(\lambda\) is used to control the dissimilarity of features at the various stages of training. It can be small at the beginning of training to facilitate better morphological modeling, then increased to learn domain invariant features later on.

\paragraph{Training Process} For each of the training batches, we populate half of the batch with samples from the morphologically labeled data, and the other half with the unlabeled data. The model calculates the morphological tagging loss for the first half, and the discriminator loss with the other, and optimizes for both jointly.

\section{Experiments and Results}
In this section we first discuss the datasets that we use, along with the experimental setup for the various experiments. We then discuss the results of the different models, using the full training datasets, and a learning curve over the \textsc{Egy} dataset, to simulate low-resource settings.

\subsection{Data}

\paragraph{Labeled Data} For \textsc{MSA} we use the Penn Arabic Treebank (PATB parts 1, 2, and 3) \cite{Maamouri:2004}.
For \textsc{Egy}, we use the ARZ Treebank (ARZTB) annotated corpus from the Linguistic Data Consortium (LDC), parts 1, 2, 3, 4, and 5 \cite{ARZMAG:2012}. The annotation process and features are similar to those of \textsc{MSA}. We follow the data splits recommended by \newcite{diab2013ldc} for training, development, and testing, for both MSA and \textsc{Egy}.
Table~\ref{dataset} shows the data sizes. Throughout the different experiments in this paper, the \textsc{Dev Test} dataset is used during the system development to assess design choices. The \textsc{Blind Test} dataset is used after finalizing the architecture, to evaluate the system and present the overall results. We use Alif/Ya and Hamza normalization, and we remove all diacritics (besides for lemmas and diacritized forms) for all variants. 

\begin{table}[h]
\centering
\begin{footnotesize}
\begin{tabular}{|c|c|c|c|}
\hline
 & \textbf{\textsc{Train}} & \textbf{\textsc{Dev Test}} & \textbf{\textsc{Blind Test}} \\ \hline \hline
\textbf{\textsc{MSA}} & 503K & 63K & 63K \\ \hline
\textbf{\textsc{Egy}} & 134K & 21K & 20K \\ \hline
\end{tabular}
\end{footnotesize}
\caption{Word count statistics for MSA and \textsc{Egy}.}
\label{dataset}
\end{table}

The morphological analyzers that we use include SAMA \cite{SAMA31} for \textsc{MSA}, and a combination of SAMA, CALIMA \cite{CALIMA:2012}, and ADAM \cite{salloum2014adam} for \textsc{Egy}, as used in the MADAMIRA \cite{MADAMIRA:2014} system.

\paragraph{Unlabeled Data} The pretrained word embeddings for MSA are trained using the LDC's Gigaword corpus \cite{LDC:Gigaword-5}. For \textsc{Egy} we use about 410 million words of the Broad Operational Language Translation (BOLT) Arabic Forum Discussions \cite{BOLT:2018}. We use the MADAR corpus \cite{Bouamor-etal-2018} as the seed dictionary for embedding space mapping.
We use the \textsc{Egy} data from the work by \newcite{zbib2012machine} as the unlabeled corpus for \textsc{Egy}. 

\subsection{Experimental Setup}

\paragraph{Tagging Architecture} We use two hidden layers of size 800 for the Bi-LSTM network (two for each direction), and a dropout wrapper with keep probability of 0.7, and peephole connections. We use Adam optimizer \cite{Kingma-arxiv-2014} with a learning rate of 0.0005, and cross-entropy cost function. We run the various models for 70 epochs (fixed number of epoch since we use dropout). The LSTM character embedding architecture uses two LSTM layers of size 100, and embedding size 50. 
We use Word2Vec \cite{Mikolov-NIPS-2013} to train the word embeddings. The embedding size is 250, and the embedding window is of size two.

\paragraph{Adversarial Adaptation} For the adversarial adaptation experiments we first observed that the average sentence length in the unlabeled \textsc{Egy} dataset is very short compared to the MSA dataset (5 words per sentence for the unlabeled dataset, and 31 words per sentence for MSA). The difference in sentence length results in the unlabeled \textsc{Egy} dataset being four times the number of batches compared to MSA, for the same number of tokens, and the model was not converging. We therefore use a minimum sentence length of 14 words for the unlabeled dataset, which results in about 9K sentences ($\sim$185K tokens).
We also found that a constant \(\lambda\) value of one performed better than scheduling the value starting from zero.

\paragraph{Metrics} The evaluation metrics we use include:

\begin{itemize}
\item POS accuracy (\textsc{POS}): The accuracy of the POS tags, of a tagset comprised of 36 tags \cite{Habash_13_NAACL}.
\item The non-lexicalized morphological features accuracy (\textsc{Feats}): The accuracy of the combined 14 closed morphological features.
\item Lemmatization accuracy (\textsc{Lemma}): The accuracy of the fully diacritized lemma.
\item Diacritized forms accuracy (\textsc{Diac}): The accuracy of the diacritized form of the words.
\item Full Analysis Accuracy (\textsc{Full}): The overall accuracy over the full analysis; \mbox{\textsc{Feats} (including POS)+\textsc{Lemma}+\textsc{Diac}}, which is the strictest evaluation approach.
\end{itemize}

\paragraph{Baselines}

The baselines are based on separate models for the different features. The first baseline is MADAMIRA \cite{MADAMIRA:2014}, which is a popular morphological disambiguation tool for Arabic. MADAMIRA uses SVM taggers for the different non-lexical features, and n-gram language models for the lemmas and diacritized forms. We also use the neural extensions of MADAMIRA \cite{Zalmout-2017-EMNLP,Zalmout_2018_NAACL}, which are based on a similar architecture, but use LSTM taggers instead of the SVM models, and LSTM-based language models instead of the n-gram models. 

\begin{table*}[h]
\centering
\setlength{\tabcolsep}{3.0pt}
\def\arraystretch{1.02}
\begin{footnotesize}
\begin{tabular}{|l|c|c|c|c|c||c|c|c|c|c|}
\hline
 \multirow{2}{*}{\textbf{\textsc{Model}}}& \multicolumn{5}{c||}{\textbf{\textsc{Dev Test}}} & \multicolumn{5}{c|}{\textbf{\textsc{Blind Test}}} \\ \cline{2-11}
 & \textsc{Full} & \textsc{Feats} & \textsc{Diac} & \textsc{Lex} & \textsc{POS} & \textsc{Full} & \textsc{Feats} & \textsc{Diac} & \textsc{Lex} & \textsc{POS} \\ \hline \hline
MADAMIRA\textsubscript{MSA}  \cite{MADAMIRA:2014} & 85.6 & 87.1 & 87.7 & 96.3 & 97.1 & 85.6 & 87.3 & 87.6 & 96.3 & 97.0 \\ \hline
MSA\textsubscript{separate features} \cite{Zalmout-2017-EMNLP} & 90.4 & 92.3 & 92.4 & 96.9 & 97.9 & 90.1 & 92.3 & 92.1 & 96.6 & 97.8 \\ \hline
MSA\textsubscript{MTL:MSA} & 90.8 & 92.7 & 92.7 & 96.9 & 97.9 & 90.8 & 93.0 & 92.5 & 96.7 & 97.9 \\ \hline
MSA\textsubscript{MSA+\textsc{Egy}} merged training datasets & 90.1 & 91.9 & 91.8 & 96.9 & 97.8 & 89.8 & 92.0 & 91.4 & 96.5 & 97.7  \\ \hline
MSA\textsubscript{MTL:MSA+\textsc{Egy}}  mapped embedding spaces & 90.6 & 92.5 & 92.4 & 96.8 & 97.8 & 90.3 & 92.5 & 91.9 & 96.5 & 97.7 \\ \hline
MSA\textsubscript{MTL:MSA+\textsc{Egy}}  merged embedding corpora & 91.1 & 93.0 & 92.9 & 96.9 & 97.9 & 91.0 & 93.2 & 92.6 & 96.7 & \textbf{98.0} \\ \hline
MSA\textsubscript{MTL:MSA+\textsc{Egy}}  separate embedding spaces & 91.2 & 93.1 & 92.9 & \textbf{97.0} & \textbf{98.0} & 91.1 & 93.3 & 92.7 & 96.7 & \textbf{98.0} \\ \hline
\verb| |+ shared output layers per feature & \textbf{91.4} & \textbf{93.3} & \textbf{93.1} & \textbf{97.0} & \textbf{98.0} & \textbf{91.2} & \textbf{93.4} & \textbf{92.8} & \textbf{96.8} & \textbf{98.0} \\ \hline
\verb|   |+ shared character embeddings & 91.2 & 93.1 & 93.0 & \textbf{97.0} & \textbf{98.0} & 91.1 & 93.3 & 92.7 & 96.7 & 97.9 \\ \hline 
MSA\textsubscript{MTL:MSA+\textsc{Egy}} Adversarial Dialect Adaptation* & 91.3 & 93.2 & 93.0 & \textbf{97.0} & \textbf{98.0} & \textbf{91.2} & 93.3 & \textbf{92.8} & 96.7 & 97.9 \\ \hline \hline
MADAMIRA\textsubscript{\textsc{Egy}}  \cite{MADAMIRA:2014} & 76.2 & 86.7 & 82.4 & 86.4 & 91.7 & 77.3 & 86.9 & 83.3 & 87.3 & 91.8 \\ \hline
\textsc{Egy}\textsubscript{separate features} \cite{Zalmout_2018_NAACL} & 77.0 & 88.8 & 82.9 & 87.6 & 92.9 & 78.0 & 88.8 & 83.6 & 87.8 & 93.3 \\ \hline
\textsc{Egy}\textsubscript{MTL:\textsc{Egy}} & 77.2 & 88.8 & 82.9 & 87.6 & 93.1 & 78.1 & 88.8 & 83.5 & 88.0 & 93.4 \\ \hline
\textsc{Egy}\textsubscript{MSA+\textsc{Egy}} merged training datasets & 77.1 & 88.9 & 82.7 & 87.6 & \textbf{93.5} & 78.2 & 89.0 & 83.5 & 88.0 & \textbf{93.8} \\ \hline
\textsc{Egy}\textsubscript{MTL:MSA+\textsc{Egy}}  mapped embedding spaces & 76.7 & 88.3 & 82.6 & 87.3 & 92.7 & 78.0 & 88.6 & 83.3 & 87.8 & 93.3 \\ \hline
\textsc{Egy}\textsubscript{MTL:MSA+\textsc{Egy}} merged embedding corpora & 77.2 & 89.0 & 82.9 & \textbf{87.7} & 93.1 & 78.1 & 88.9 & 83.5 & 88.0 & 93.5 \\ \hline
\textsc{Egy}\textsubscript{MTL:MSA+\textsc{Egy}}  separate embedding spaces & 77.3 & 89.0 & 83.0 & \textbf{87.7} & 93.1 & 78.4 & 89.2 & 83.7 & 88.0 & 93.6 \\ \hline
\verb| |+ shared output layers per feature & 77.4 & 89.1 & 83.0 & \textbf{87.7} & 93.2 & 78.5 & 89.3 & \textbf{83.8} & 88.0 & 93.7 \\ \hline
\verb|   |+ shared character embeddings & 77.3 & 89.0 & 82.9 & \textbf{87.7} & 93.2 & 78.2 & 89.1 & 83.6 & \textbf{88.1} & 93.7 \\ \hline
\textsc{Egy}\textsubscript{MTL:MSA+\textsc{Egy}} Adversarial Dialect Adaptation* & \textbf{77.5} & \textbf{89.3} & \textbf{83.1} & \textbf{87.7} & 93.3 & \textbf{78.6} & \textbf{89.4} & \textbf{83.8} & \textbf{88.1} & \textbf{93.8} \\ \hline
\end{tabular}
\end{footnotesize}
\caption{Disambiguation results for joint MSA and \textsc{Egy} modeling. MTL is Multitask Learning. *Best adversarial result was with merged embedding spaces.}
\label{msa-egy-results}
\end{table*}

\subsection{Results}

To evaluate the performance of the knowledge-transfer scheme, we present the results in two parts. The first presents the results for the full MSA and \textsc{Egy} datasets, evaluating the accuracy of the various architecture configurations. We then present the results of a learning curve over the size of the \textsc{Egy} training dataset, modeling various degrees of low-resource performance. The goal is to assess the multitask learning and adversarial training models in particular, and the degree of knowledge-transfer, which should be more helpful when the size of the \textsc{Egy} training data is lower.

\subsubsection{Joint Morphological Modeling}

Table~\ref{msa-egy-results} shows the results of the joint modeling of MSA and \textsc{Egy}. Based on the results, we make the following observations:

\paragraph{Multi-Feature Modeling} The results for the multi-feature models show consistent and significant improvement compared to the separate models for each feature, especially for MSA.  This supports the assumption that multi-feature modeling can identify more complex patterns involving multiple features, that separate models cannot.

\paragraph{Cross-Dialectal Modeling: Merged Training Data vs Multitask Learning} For the cross-dialectal MSA and \textsc{Egy} models, we first experiment with merging the training datasets for both, and train a single model over the merged datasets. This model is a simple baseline for the cross-dialectal models, but imposes hard joint modeling that might lead to some knowledge loss.

The results indicate that the multitask learning architecture performs much better, especially for MSA. The accuracy for POS tagging for \textsc{Egy} in particular was higher or similar though. This is probably because POS behaves very similarly in both MSA and \textsc{Egy}, unlike other morphological features that might converge slightly. So the added MSA training samples were generally helpful.

\paragraph{Embedding Models}  Joint embedding spaces between the dialects, whether through embedding space mapping or through learning the embeddings on the combined corpus, did not perform well. Using separate embedding models (whether for word or character embeddings) for each dialect shows better accuracy.
Embedding models learn properties and morphosyntactic structures that are specific to the training data. Mapping the embedding spaces likely results in some knowledge loss.
Unlike the adversarial training model though, at which the merged embedding datasets model performed better. This is expected since the goal of adversarial training is to bring the overall feature spaces closer to learn dialect-invariant features.

\paragraph{Shared Output Layers} 
The results indicate that using shared output layers for the different dialects improves the overall accuracy. Shared output layers are more likely to learn shared morphosyntactic structures from the other dialect, thus helping both. Having separate layers wastes another joint learning potential. The shared output layers further reduce the size of the overall model.

\paragraph{Adversarial Dialect Adaptation} The adversarial adaptation experiments show slightly higher results for \textsc{Egy}, but very close results to the multitask learning model for MSA. Since MSA is resource-rich it is expected that adversarial training would not be beneficial (or even hurtful), as the dialect-invariant features would hinder the full utilization of the rich MSA resources. For \textsc{Egy}, we expect that the knowledge-transfer model would be more beneficial in lower-resource scenarios, we therefore experiment with a learning curve for the training dataset size in the next section.

\subsubsection{Modeling Training Data Scarcity}

\begin{table}[h]
\centering
\setlength{\tabcolsep}{10pt}
\def\arraystretch{1.05}
\begin{footnotesize}
\begin{tabular}{|c|c|c|c|}
\hline
\multirow{2}{*}{\textbf{\textsc{Egy Train Size}}} & \multirow{2}{*}{\textbf{\textsc{EGY}}} & \multicolumn{2}{c|}{\textbf{MSA-\textsc{EGY}}} \\ \cline{3-4} 
 &  & \textbf{MTL} & \textbf{\textsc{Adv}} \\ \hline \hline
\textbf{2K (1.5\%)} & 29.7 & 61.9 & \textbf{71.1} \\ \hline
\textbf{8K (6\%)} & 62.5 & 73.5 & \textbf{78.3} \\ \hline
\textbf{16K (12\%)} & 74.7 & 78.1 & \textbf{81.5} \\ \hline
\textbf{33K (25\%)} & 80.7 & 81.6 & \textbf{83.5} \\ \hline
\textbf{67K (50\%)} & 83.3 & 82.0 & \textbf{84.0} \\ \hline
\textbf{134K (100\%)} & 84.5 & 85.4 & \textbf{85.6} \\ \hline
\end{tabular}
\end{footnotesize}
\caption{The results (\textsc{Feats}) of the learning curve over the \textsc{Egy} training dataset, for the \textsc{Egy} dataset alone, multitask learning (MTL), and the adversarial training (\textsc{Adv}). We do not use morphological analyzers here, so the results are not comparable to Table \ref{msa-egy-results}.} 
\label{learning-curve}
\end{table}

Knowledge-transfer schemes are more valuable in low-resource settings for the target language. To simulate the behavior of the multitask and adversarial learning architectures in such setting, we train the model using fractions of the \textsc{Egy} training data. We reduce the training dataset size by a factor of two each time. We then simulate extreme scarcity, having only 2K \textsc{Egy}  annotated tokens.

Low-resource dialects will have very limited or no morphological analyzers, so we also simulate the lack of morphological analyzers for \textsc{Egy}. Since we are not using an \textsc{Egy} morphological analyzer, we evaluate the models on the set of non-lexicalized and clitics features only, without the diacritized forms and lemmas. We also do not perform an explicit disambiguation step through analysis ranking, and we evaluate on the combined morphological tags directly for each word. 

Table \ref{learning-curve} shows the results. Multitask learning with MSA consistently outperforms the models that use \textsc{Egy} data only. The accuracy almost doubles in the 2K model. We also notice that the accuracy gap increases as the  \textsc{Egy} training dataset size decreases, highlighting the importance of joint modeling with MSA in low-resource DA settings.
The adversarial adaptation results in the learning curve further show a significant increase in accuracy with decreasing training data size, compared to the multitask learning results. The model seems to be facilitating more efficient knowledge-transfer, especially for the lower-resource \textsc{Egy} experiments. We can also observe that for the extreme low-resource setting, we can double the accuracy through adversarial multitask learning, achieving about 58\% relative error reduction.

The results also indicate that with only 2K \textsc{Egy} annotated tokens, and with adversarial multitask learning with MSA, we can achieve almost the same accuracy as 16K tokens using \textsc{Egy} only. This is a significant result, especially when commissioning new annotation tasks for other dialects.

\paragraph{Error Analysis}

We investigated the results in the learning curve to understand the specific areas of improvement with multitask learning and adversarial training. We calculated the accuracies of each of the features, for both models, and across all the dataset sizes. We observed that the POS and Gender features benefited the most of the joint modeling techniques. Whereas features like Mood and Voice benefited the least.  
This is probably due to the relatively similar linguistic behavior for POS and Gender in both MSA and \textsc{Egy}, unlike Mood or Voice, which are less relevant to DA, and can be somewhat inconsistent with MSA. The improvement was consistent for both approaches, and across the training data sizes, with POS having almost 61\% relative error reduction in the 2K dataset with adversarial training, and Mood (the least improving feature) of about 8\%. And 8\% for POS, and 0\% for Mood, in the full size dataset.

\section{Conclusions and Future Work}
In this paper we presented a model for joint morphological modeling of the features in morphologically rich dialectal variants. We also presented several extensions for cross-dialectal modeling. We showed that having separate embedding models, but shared output layers, performs the best. Joint modeling for the features within each dialect performs consistently better than having separate models, and joint cross-dialectal modeling performs better than dialect-specific models. We also used  adversarial training to facilitate a knowledge-transfer scheme, providing the best result for \textsc{Egy}, especially in lower-resource cases. Our models result in state-of-the-art results for both MSA, and \textsc{Egy}. 
Future work includes joint and cross-dialectal lemmatization models, in addition to further extension to other dialects.

 \paragraph{Acknowledgment}
The first author was supported by the New York University Abu Dhabi Global PhD Student Fellowship program. The support and resources from the High Performance Computing Center at New York University Abu Dhabi are also gratefully acknowledged.

\bibliographystyle{acl_natbib}

\end{document}